\documentclass[11pt,a4paper]{article}
\usepackage[hyperref]{emnlp2018}
\usepackage{times}
\usepackage{latexsym}
\usepackage{graphicx}
\usepackage{forest}
\usepackage{multirow}
\usepackage{tabularx}
\usepackage{soul}
\usepackage{algorithm,setspace}
\usepackage{amsmath}
\usepackage[noend]{algpseudocode}

\usepackage{url}

\aclfinalcopy 

\title{Affordance Extraction and Inference \\ based on Semantic Role Labeling}

\author{Daniel Loureiro, Al\'ipio M\'ario Jorge \\
  LIAAD - INESC TEC \\
  Faculty of Sciences - University of Porto, Portugal \\
  {\tt dloureiro@fc.up.pt, amjorge@fc.up.pt}}

\date{}

\begin{document}
\maketitle
\begin{abstract}
	Common-sense reasoning is becoming increasingly important for the advancement of Natural Language Processing. While word embeddings have been very successful, they cannot explain which aspects of `coffee' and `tea' make them similar, or how they could be related to `shop'. In this paper, we propose an explicit word representation that builds upon the Distributional Hypothesis to represent meaning from semantic roles, and allow inference of relations from their meshing, as supported by the affordance-based Indexical Hypothesis. We find that our model improves the state-of-the-art on unsupervised word similarity tasks while allowing for direct inference of new relations from the same vector space.
\end{abstract}

\section{Introduction}

The word representations used more recently in Natural Language Processing (NLP) have been based on the Distributional Hypothesis (DH) \cite{harris1954} | ``words that occur in the same contexts tend to have similar meanings''. This simple idea has led to the development of powerful word embedding models, starting with Latent Semantic Analysis (LSA) \cite{Landauer1997AST} and later, the popular word2vec \cite{Mikolov2013LinguisticRI} and GloVe \cite{Pennington2014GloveGV} models. Although, effective at quantifying the similarity between words (and phrases) such as `tea' and `coffee', they cannot relate that judgement to the fact that both can be sold, for instance. Furthermore, current representations can't inform us about possible relations between words occurring in mostly distinct contexts, such as using a `newspaper' to cover a `face'. While there have been substantial improvements to word embedding models over the years, these shortcomings have endured \cite{CamachoCollados2018FromWT}.

\begin{table}[htb]
\centering
\scalebox{0.9}{
\begin{tabular}{|c|c|}
\hline
\textbf{Word Pairs}  & \textbf{Affordances}                \\
\textbf{\small{(w\textsubscript{1}, w\textsubscript{2})}}    & \textbf{\small{(w\textsubscript{1} as ARG0, w\textsubscript{2} as ARG1)}} \\ \hline
shop, tea            & sell, import, cure                  \\ \hline
doctor, patient      & diagnose, prescribe, treat          \\ \hline
newspaper, face      & cover, expose, poke                 \\ \hline
man, cup             & drink, pour, spill                  \\ \hline
\end{tabular}}
\caption{Results from affordance meshing (coordination) using automatically labelled semantic roles.}
\label{table:affordances}
\end{table}

\citet{Glenberg2000SymbolGA} identified these issues soon after LSA was introduced, and cautioned that high-dimensional word representations, such as those based on the DH, lack the necessary grounding to be proper semantic analogues. Instead, Glenberg proposed the Indexical Hypothesis (IH) which supports that meaning is constructed by (a) indexing words and phrases to real objects or perceptual, analog symbols; (b) deriving affordances from the objects and symbols; and (c) meshing the affordances under the guidance of syntax. Following \citet{Glenberg2000SymbolGA}, this work considers an object's affordances as its possibilities for action constrained by its context, including actions which may not be directly perceived, which differs slightly from \citet{Gibson79}'s original definition. Even though the language grounding advocated by the IH is beyond the reach of NLP by itself, we believe that its representation of meaning through affordances can still be captured to a useful extent. 



Our contribution\footnote{Code, data and demo: \href{https://a2avecs.github.io}{https://a2avecs.github.io}} is a word-level representation that allows for the affordance correspondence and meshing supported by the IH. These affordances are approximated from occurrences of semantic roles in corpora through an adaptation of models based on the DH. Our work is motivated by two observations: (1) a pressing need to integrate common-sense knowledge in NLP models and (2) recent improvements to Semantic Role Labeling (SRL) have made affordance extraction from raw corpora sufficiently reliable. We find that our model (A2Avecs) performs competitively on word similarity tasks while enabling novel `who-does-what-to-whom' style inferences (Table \ref{table:affordances}).

\section{Related Work}

This work is closely related to the research area of selectional preferences, where the goal is to predict the likelihood of a verb taking a certain argument for a particular role (e.g. likelihood of \textit{man} being an \textit{agent} of \textit{drive}). Most notably, \citet{Erk2010AFC} proposed a distributional model of selectional preferences that used SRL annotations as a primary set of verb-role-arguments from which to generalize using word representations based on the DH and several word similarity metrics. Progress in selectional preferences is usually measured through correlations with human thematic fit judgements and, more recently, neural approaches \cite{Cruys2014ANN, Tilk2016EventPM} obtained state-of-the-art results.

While this work shares some of these same elements (i.e. SRL and word embeddings), they are used to predict potential affordances instead of selectional preferences. Consequently, our representations are designed to enable the meshing proposed by the IH, allowing us to infer affordances that would not be likely under a selectional preference learning scheme (e.g. \textit{newspaper-cover-face} from Table \ref{table:affordances}). Additionally, this work is concerned with showing that information derived from SRL is complementary to information derived from DH methods, and thus focuses its evaluation on tasks related to lexical similarity rather than thematic fit correlations.

\begin{figure}[tb]
\centering
\includegraphics[width=0.4\textwidth]{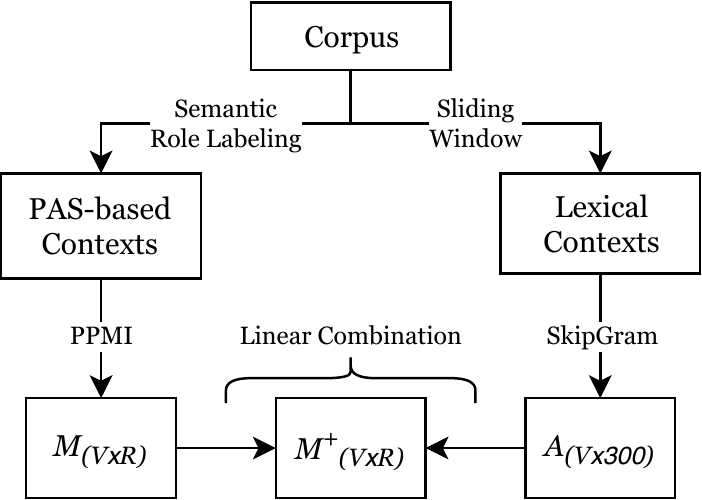}
\caption{Outline of model pipeline.}
\label{diagram:outline}
\end{figure}

\section{Method}

Our word representations are modelled using Predicate-Argument Structures (PASs). These structures are obtained through SRL of raw corpora, and used to populate a sparse word/context co-occurrence matrix $W$ where roles serve as contexts (features), and argument spans serve as the co-occurrence windows. The roles are predicates specified by argument type (e.g. eat$\vert$ARG0) and used in place of affordances. See Table \ref{table:contexts} for a comparison of this context definition with the traditional lexical definition.

\begin{table}[hbp]
\centering
\scalebox{0.85}{
\begin{tabular}{l|l|l}
                   & \textbf{Context} & \textbf{Words}             \\ \hline
\multirow{3}{*}{\rotatebox{90}{\textbf{Role}}} & drinks$\vert$ARG0      & John                       \\
                   & drinks$\vert$ARG1      & red, wine                  \\
                   & drinks$\vert$ARGM-MNR  & slowly                     \\ \hline
\multirow{5}{*}{\rotatebox{90}{\textbf{Adjacency}}} & John             & drinks, red                \\
                   & drinks           & John, red, wine            \\
                   & red              & John, drinks, wine, slowly \\
                   & wine             & drinks, red, slowly        \\
                   & slowly           & red, wine                 
\end{tabular}}
\caption{Different context definitions applied to the sentence `John drinks red wine slowly'. Top: Our proposed definition; Bottom: Lexical adjacency definition (with window size of 2).}
\label{table:contexts}
\end{table}

After computing our co-occurrence matrix we follow-up with the additional steps employed by traditional bag-of-words models. We use Positive Pointwise Mutual Information (PPMI) to improve co-occurrence statistics, as used successfully by \citet{Bullinaria2007ExtractingSR, Levy2014LinguisticRI}, and maintain explicit high-dimensional representations in order to preserve the context information required for affordance meshing. Previous works, such as \citet{Levy2014DependencyBasedWE} and \citet{Stanovsky2015OpenIA}, have also produced word representations from syntactic context definitions (dependency parse trees and open information extraction, respectively) but have opted for following-up with the word2vec's SkipGram (SG) model, presumably influenced by a much higher number of contexts in their approaches.


We reduce the sparsity of our explicit PPMI matrix by linear combination
and interpolation of semantically related vectors. The semantic relatedness is  obtained from the cosine similarity of SG vectors. As evidenced by \citet{Baroni2014DontCP}, SG seem best suited for estimating relatedness (or association). These steps are further described in remainder of this section (See Fig. \ref{diagram:outline}).


\subsection{Extracting PASs}

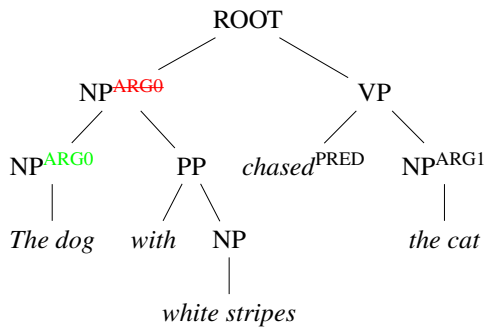
\begin{figure}[hbp]
\centering
\scalebox{0.9}{
\begin{forest}
  [ROOT
    [NP\textsuperscript{\color{red}\st{ARG0}}
     [NP\textsuperscript{\color{green}ARG0} [\textit{The dog}]]
     [PP
       [\textit{with}]
       [NP [\textit{white stripes}]]
     ]
    ]
    [VP
      [\textit{chased}\textsuperscript{PRED}]
      [NP\textsuperscript{ARG1} [\textit{the cat}]]
    ]
  ]
\end{forest}}
\caption{Parse tree for the sentence `The dog with white stripes chased the cat.'. The label for ARG0 is repositioned to the smaller subtree.}
\label{figure:tree}
\end{figure}

We use the AllenNLP \cite{Gardner2017AllenNLP} implementation of \citet{He2017DeepSR} state-of-the art SRL to extract PASs from an English Wikipedia dump from April  2010 (1B words). Since the automatic identification of predicates by an end-to-end SRL may produce erroneous results, we ensure that these predicates are valid by restricting them to the set of verbs tagged on the Brown corpus \cite{francis1979brown}. We also use the spaCy parser \cite{spacy2} to reduce each argument phrase to its head noun phrase, reducing the dilution of the more relevant noun and predicate co-occurrence statistics (See Fig. \ref{figure:tree}). Additionally, we lemmatize the predicates (verbs) to their root form using WordNet's Morphy Lemmatizer \cite{Miller1992WORDNETAL}. Finally, we trim the vocabulary size and the number of roles by discarding those which occur less than 100 times, and consider only core and adjunct argument types. The result is a set $C$ of observed contexts, such as $<$chase$\vert$ARG1, (the, cat)$>$, used to populate $W$.

\subsection{Argument-specific PPMI}

The authors of PropBank \cite{Kingsbury2002FromTT}, which provides the annotations used for learning SRL, state that arguments are predicate-specific. Still, they also acknowledge that there are some trends in the argument assignments. For instance, the most consistent trends are that ARG0 is usually the agent, and ARG1 is the direct object or theme of the relation. This realisation leads us to adapting the PPMI measure to better account for the correlations between roles of the same argument types.
Thus, we segment $C$ by argument type, and apply PPMI independently considering each $C_{ARG}$, such that for each $W_{w,p}$:
$$PPMI(w, r) = max(PMI(w, r), 0))$$
{\small{$$PMI(w, r) = log \frac{f(w, r)}{f(w)f(r)} = log \frac{\#(w, r)\cdot |C_{ARG}|}{\#(w)\cdot \#(r)}$$}}
where $w$ is a word from the vocabulary $V$, $r$ is a role (context) from the set $R$ of the same argument type as $C_{ARG}$, and $f$ is the probability function. The resulting matrix $M=PPMI(W)$ maintains the  dimensions $W$ and is slightly sparser.

\subsection{Leveraging Association}

\begin{table}[t!]
\scalebox{0.95}{
\centering
\begin{minipage}{0.5\textwidth}
\begin{tabular}{cccc|c|c}
\textbf{PST} & \textbf{PMI} & \textbf{THR} & \textbf{RND} & \textbf{MEN} & \textbf{SP} \\ \hline
L &  &  &  & .249 (1) & 98.71 \\
L+H &  &  &  & .309 (47) & 99.41 \\
L+H & P &  &  & .606 (47) & 99.58 \\
L+H & A+P &  &  & .611 (47) & 99.59 \\
L+H & A+P & 0.5 &  & N/A\footnote{Failed after using too much memory.} & \small{$<41.2$} \\
L+H & A+P & 0.5 & HDR & \textbf{.687 (0)} & 98.21 \\
L+H & A+P & 0.6 & HDR & .654 (14) & 97.98 \\
L+H & A+P & 0.4 & HDR & .668 (0) & 98.77
\end{tabular}
\end{minipage}}
\caption{Sensitivity/Impact analysis for some parameters of our approach. \\\small{Legend: \textbf{PST}: Post-Processing (L: Lemmatization; H: Head noun phrase isolation); \textbf{PMI}: PMI Variations (P: PPMI; A: Arg-specific PPMI); \textbf{THR}: Similarity threshold (tested values); \textbf{RND}: Rounding (HDR: Half down rounding); \textbf{MEN}: MEN-3K task (Spearman correlation, \#OOV failures); \textbf{SP}: Sparsity (percentage of zero values on a 155Kx18K matrix).}}
\label{table:component}
\end{table}

The constraints imposed by SRL yield a very reduced number of PAS-based contexts that can be extracted from a corpus, in comparison to lexical adjacency-based contexts. Moreover, the post-processing steps we perform, while otherwise beneficial (see Table \ref{table:component}), further trim this information. To mitigate this issue, we also compute an embedding matrix $A$ (see Section 3 for parameters), using the state-of-the-art lexical-based SG model of \citet{Bojanowski2017EnrichingWV}, and use those embeddings to obtain similarity values that can be used to interpolate missing values in $M$, through weighted linear combination. This way, existing vectors are re-computed as:

{$$\vec{v}_w = \frac{\vec{v}_1*\alpha_1+...+\vec{v}_n*\alpha_n}{\sum_{i=1}^{n}\alpha_i}$$}


\vspace{2cm}

with $\alpha_i$ defined as:

\[
    \alpha_i = 
\begin{cases}
    \frac{A_{\vec{v}_w} \cdot A_{\vec{v}_i}} {\parallel A_{\vec{v}_w} \parallel \parallel A_{\vec{v}_i} \parallel} = cos_A(\vec{v}_w, \vec{v}_i),\\ \hspace{23mm} \text{if}\hspace{1mm}cos_A(\vec{v}_w, \vec{v}_i) > 0.5\\
    0,\hspace{2mm}\text{otherwise}
\end{cases}
\]

where $cos_A$ corresponds to the cosine similarity in the SG representations.

\medskip

The similarity threshold is tested on a few natural choices ($0.5 \pm 0.1$) and validated from results on a single word similarity task (see Table \ref{table:component}). This approach is also used to define representations for words that are out-of-vocabulary (OOV) for $M$, but can be interpolated from related representations, similarly to \citet{Zhao2015LearningTM}. In conjunction with the interpolation, we apply half down rounding to the vectors, before and after re-computing them, so that our representations remain efficiently sparse while benefitting from improved performance. Finally, we apply a quadratic transformation to enlarge the influence of meaningful co-occurrences, obtaining \hspace{4cm} $M^+ = interpolate(M, A)^2$.

\subsection{Inferring Relations}

The examples shown in Table \ref{table:affordances} are easily obtained with our model through a simple procedure (see Algorithm \ref{algorithm:inference}) that matches different arguments of the same predicates. As was the case with Arg-specific PPMI, this procedure is made possible by the fact that a significant portion of argument assignments remain consistent across predicates.

\begin{algorithm}[h]
\setstretch{1.25}
\small
\caption{Affordance Meshing Algorithm}
\label{algorithm:inference}
\begin{algorithmic}[1]
\Procedure{INFERENCE}{$M^+, w_1, w_2, a_1, a_2$}
\State $relations \gets []$
\State $\vec{v}_1, \vec{v}_2\gets get\_vec(w_1, M^+), get\_vec(w_2, M^+)$
\For{$f_1 \in features(\vec{v}_1) \wedge arg(f_1) = a_1$}
\For{$f_2 \in features(\vec{v}_2) \wedge arg(f_2) = a_2$}
\If{$pred(f_1) = pred(f_2)$}
\State {$relations.add((f_1*f_2, pred(f_1)))$}
\EndIf
\EndFor
\EndFor
\State \textbf{return} $sorted(relations)$
\EndProcedure
\end{algorithmic}
\end{algorithm}

\section{Evaluation and Experiments}

The A2Avecs model introduced in this paper is used to generate 155,183 word vectors of 18,179 affordance dimensions. This section compares our model with lexical-based models (word2vec \cite{Mikolov2013LinguisticRI}, GloVe \cite{Pennington2014GloveGV} and fastText \cite{Bojanowski2017EnrichingWV}) and other syntactic-based models (Deps \cite{Levy2014DependencyBasedWE} and OpenIE \cite{Stanovsky2015OpenIA}). We're using Deps and OpenIE embeddings that the respective authors trained on a Wikipedia corpus and distributed online. Lexical models were trained using the same parameters, wherever applicable: Wikipedia corpus from April 2010 (same as mentioned in section 2.1); minimum word frequency of 100; window size of 2; 300 vector dimensions; 15 negative samples; and ngrams sized from 3 to 6 characters.

We also show that our approach can make use of high-quality pretrained embeddings. We experiment with a fastText model pretrained on 600B tokens, referred to as `fastText 600B' in contrast with the fastText model trained on Wikipedia.

\subsection{Model Introspection}

The explicit nature of the representations produced by our model makes them directly interpretable, similarly to other sparse representations such as \citet{Faruqui2015NondistributionalWV}. The examples presented at Table \ref{table:introspection} demonstrate the relational capacity of our model, beyond associating meaningful predicates. In this introspection we highlight the top role contexts for a set of words, inspired by \cite{Levy2014DependencyBasedWE} which presented the top syntactic context for the same words, and note that this introspection produces results that should correspond to \citet{Erk2010AFC}'s inverse selectional preferences. 

Our online demonstration provides access to additional introspection queries, such as top words for given affordances, or which affordances are most distinguishable between a pair of words (determined by absolute difference).

\begin{table}[h]
\centering
\scalebox{0.68}{
\begin{tabular}{lll}
\hline
\textbf{batman}                                                                                                         & \textbf{hogwarts}                                                                                                      & \textbf{turing}                                                                                                    \\ \hline
\begin{tabular}[c]{@{}l@{}}foil$\vert$ARG0\\ flirt$\vert$ARGM-MNR\\ apprehend$\vert$ARG0\\ subdue$\vert$ARG0\\ rescue$\vert$ARGM-DIR\end{tabular}     & \begin{tabular}[c]{@{}l@{}}ambush$\vert$ARGM-MNR\\ rock$\vert$ARGM-LOC\\ express$\vert$ARG0\\ prevent$\vert$ARGM-LOC\\ expel$\vert$ARG2\end{tabular} & \begin{tabular}[c]{@{}l@{}}travel$\vert$ARGM-TMP\\ pass$\vert$ARGM-ADV\\ solve$\vert$ARG0\\ simulate$\vert$ARG1\\ prove$\vert$ARG1\end{tabular}  \\ \hline
\textbf{florida}                                                                                                        & \textbf{object-oriented}                                                                                               & \textbf{dancing}                                                                                                   \\ \hline
\begin{tabular}[c]{@{}l@{}}base$\vert$ARGM-MNR\\ vacation$\vert$ARG1\\ reside$\vert$ARGM-DIS\\ fort$\vert$ARG1\\ vacation$\vert$ARGM-LOC\end{tabular} & \begin{tabular}[c]{@{}l@{}}define$\vert$ARG1\\ define$\vert$ARG0\\ use$\vert$ARG1\\ implement$\vert$ARG1\\ express$\vert$ARGM-MNR\end{tabular}       & \begin{tabular}[c]{@{}l@{}}dance$\vert$ARG0\\ dance$\vert$ARGM-MNR\\ dance$\vert$ARGM-LOC\\ dance$\vert$ARGM-ADV\\ dance$\vert$ARG1\end{tabular} \\ \hline
\end{tabular}}
\caption{Words and their top role contexts. Using the same words from the introspection of \cite{Levy2014DependencyBasedWE} to clarify the difference in the representations of both approaches.}
\label{table:introspection}
\end{table}

\begin{table*}[htb]
\centering
\scalebox{0.84}{
\begin{tabular}{cccccclc}
\hline
\multicolumn{1}{|c|}{\textbf{Context}} & \multicolumn{1}{c|}{\textbf{Model}} & \textbf{SL-666} & \textbf{SL-999} & \textbf{WS-SIM} & \textbf{WS-ALL} & \textbf{MEN} & \multicolumn{1}{c|}{\textbf{RG-65}} \\ \hline
\multicolumn{1}{|c|}{\multirow{3}{*}{Lexical}} & \multicolumn{1}{c|}{word2vec} & .426 & .414 & .762 & .672 & .721 & \multicolumn{1}{c|}{.793} \\
\multicolumn{1}{|c|}{} & \multicolumn{1}{c|}{GloVe} & .333 & .325 & .637 & .535 & .636 & \multicolumn{1}{c|}{.601} \\
\multicolumn{1}{|c|}{} & \multicolumn{1}{c|}{fastText \small{($A$)}} & .426 & .419 & \textbf{.779} & \textbf{.702} & \textbf{.751} & \multicolumn{1}{c|}{.799} \\ \hline
\multicolumn{1}{|c|}{\multirow{4}{*}{Syntactic}} & \multicolumn{1}{c|}{Deps} & \textbf{.475} & \textbf{.446} & .758 & .629 & .606 & \multicolumn{1}{c|}{.765} \\
\multicolumn{1}{|c|}{} & \multicolumn{1}{c|}{Open IE} & .397 & .390 & .746 & .696 & .281 & \multicolumn{1}{c|}{.801} \\
\multicolumn{1}{|c|}{} & \multicolumn{1}{c|}{A2Avecs \small{($M^+$)}} & .461 & .412 & .734 & .577 & .687 & \multicolumn{1}{c|}{\textbf{.802}} \\
\multicolumn{1}{|c|}{} & \multicolumn{1}{c|}{A2Avecs \small{(SVD($M^+$))}} & .436 & .386 & .672 & .509 & .599 & \multicolumn{1}{c|}{.789} \\ \hline
 &  &  &  &  &  &  &  \\ \hline
\multicolumn{1}{|c|}{Lexical SOTA} & \multicolumn{1}{c|}{fastText 600B \small{($A$)}} & \multicolumn{1}{c|}{.523} & \multicolumn{1}{c|}{.504} & \multicolumn{1}{c|}{.839} & \multicolumn{1}{c|}{\textbf{.791}} & \multicolumn{1}{l|}{\textbf{.836}} & \multicolumn{1}{c|}{\textbf{.859}} \\ \hline
\multicolumn{1}{|c|}{Intp. w/SOTA} & \multicolumn{1}{c|}{A2Avecs ($M^+$)} & \multicolumn{1}{c|}{.513} & \multicolumn{1}{c|}{.468} & \multicolumn{1}{c|}{.780} & \multicolumn{1}{c|}{.619} & \multicolumn{1}{l|}{.744} & \multicolumn{1}{c|}{.814} \\ \hline
\multicolumn{1}{|c|}{Intp. \& Conc.} & \multicolumn{1}{c|}{A2Avecs (\small{$M^+ \parallel A$})} & \multicolumn{1}{c|}{\textbf{.540}} & \multicolumn{1}{c|}{\textbf{.521}} & \multicolumn{1}{c|}{\textbf{.846}} & \multicolumn{1}{c|}{.771} & \multicolumn{1}{l|}{.829} & \multicolumn{1}{c|}{.857} \\ \hline
\multicolumn{1}{|c|}{Deps Conc.} & \multicolumn{1}{c|}{Deps $\parallel A$} & \multicolumn{1}{c|}{.524} & \multicolumn{1}{c|}{.503} & \multicolumn{1}{c|}{.818} & \multicolumn{1}{c|}{.752} & \multicolumn{1}{l|}{.770} & \multicolumn{1}{c|}{.835} \\ \hline
\end{tabular}}
\caption{Spearman correlations for word similarity tasks (see \citet{Faruqui2014CommunityEA} for task descriptions). Top section shows results from training on the Wikipedia corpus exclusively. Bottom section shows results where we used SG embeddings $(A)$ trained on a larger corpus for performing interpolation and concatenation on the same set of roles used above. For comparison, we also show results for Deps concatenated with those embeddings.}
\label{table:results}
\end{table*}

\subsection{Word Similarity}

The results presented on Table \ref{table:results} show that our model can outperform lexical and syntactic models, in spite of maintaining an explicit representation. In fact, applying Singular Value Decomposition (SVD) to obtain dense 300-dimensional embeddings degrades performance. We achieve best results with the concatenation of the fastText 600B vectors with our model interpolated using those same vectors for the vocabulary $V_{M^+} \cap V_A$, after normalizing both to unit length ($L_2$). Interestingly, the same concatenation process with Deps embeddings doesn't seem as beneficial, suggesting that our representations are more complementary.

\section{Conclusion}
Our results suggest that semantic similarity can be captured in a vector space that also allows for the inference of new relations through affordance-based representations, which opens up exciting possibilities for the field. In the process, we presented more evidence to support that information obtained from SRL is complementary to information obtained from adjacency-based contexts, or even contexts based on syntactical dependencies.
We believe this work helps bridge the gap between selectional preferences and semantic plausibility, beyond frequentist generalizations based on the DH. In the near term, we expect that specific tasks such as Entity Disambiguation and Coreference can benefit from these representations. With further developments, semantic plausibility assessments should also be useful for more broad tasks such as Fact Verification and Story Understanding.

\bibliography{emnlp2018}
\bibliographystyle{acl_natbib_nourl}

\end{document}